\documentclass[sigconf]{acmart} 
\AtBeginDocument{%
  \providecommand\BibTeX{{%
    \normalfont B\kern-0.5em{\scshape i\kern-0.25em b}\kern-0.8em\TeX}}}

\copyrightyear{2020}
\acmYear{2020}
\setcopyright{acmcopyright}
\acmConference[MM '20] {28th ACM International Conference on Multimedia}{October 12--16, 2020}{Seattle, WA, USA.}
\acmBooktitle{28th ACM International Conference on Multimedia (MM '20), October 12--16, 2020, Seattle, WA, USA.}
\acmPrice{15.00}
\acmDOI{10.1145/3394171.3413825}
\acmISBN{978-1-4503-7988-5/20/10}

\usepackage{eucal}
\usepackage{multirow}
\usepackage{array}
\usepackage[normalem]{ulem}

\def\vs{\textit{vs\ }}
\def\ie{\textit{i.e.,\ }}
\def\eg{\textit{e.g.,\ }}
\def\etal{\textit{et~al.}}
\newcommand{\heading}[1]{\noindent\textbf{#1}}

\newcommand{\bc}[1]{\textcolor{blue}{#1}}
\newcommand{\rc}[1]{\textcolor{red}{#1}}

\newcommand{\para}[1]{\noindent \textbf{#1}}

\begin{document}
\fancyhead{}

\title{Towards Unsupervised Crowd Counting via Regression-Detection Bi-knowledge Transfer}

\author{Yuting Liu}
\authornote{This work was done when Yuting Liu was a visiting student at National Institute of Informatics supported by China Scholarship Council (CSC).}
\affiliation{Sichuan University}
\email{yuting.liu@stu.scu.edu.cn}

\author{Zheng Wang}
\authornote{Corresponding Author}
\affiliation{National Institute of Informatics}
\email{wangz@nii.ac.jp}

\author{Miaojing Shi}
\affiliation{King's College London}
\email{miaojing.shi@kcl.ac.uk}

\author{Shin'ichi Satoh}
\affiliation{National Institute of Informatics}
\email{satoh@nii.ac.jp}

\author{Qijun Zhao$^{\dagger}$}
\affiliation{Sichuan University}
\email{qjzhao@scu.edu.cn}

\author{Hongyu Yang}
\affiliation{Sichuan University}
\email{yanghongyu@scu.edu.cn}

\begin{abstract}
Unsupervised crowd counting is a challenging yet not largely explored task. In this paper, we explore it in a transfer learning setting where we learn to detect and count persons in an unlabeled target set by transferring bi-knowledge learnt from regression- and detection-based models in a labeled source set. The dual source knowledge of the two models is heterogeneous and complementary as they capture different modalities of the crowd distribution. We formulate the mutual transformations between the outputs of regression- and detection-based models as two scene-agnostic transformers which enable knowledge distillation between the two models. Given the regression- and detection-based models and their mutual transformers learnt in the source, we introduce an iterative self-supervised learning scheme with regression-detection bi-knowledge transfer in the target. Extensive experiments on standard crowd counting benchmarks, ShanghaiTech, UCF\_CC\_50, and UCF\_QNRF demonstrate a substantial improvement of our method over other state-of-the-arts in the transfer learning setting. 
\end{abstract}


\maketitle

\section{Introduction}

\begin{figure}[t]
\centering
    \begin{tabular}{cc}
			\includegraphics[width=0.22\textwidth]{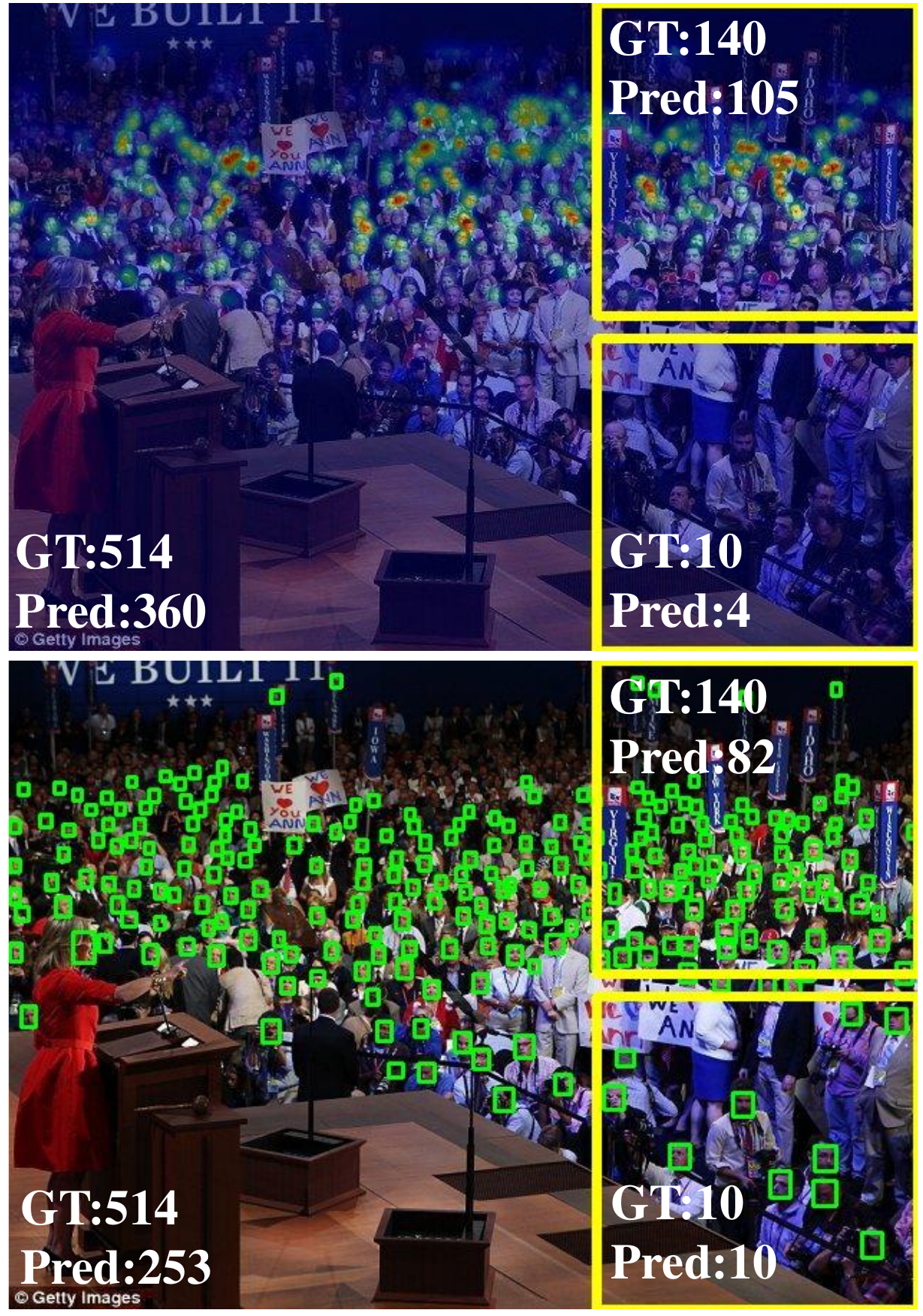}&
		    \includegraphics[width=0.23\textwidth]{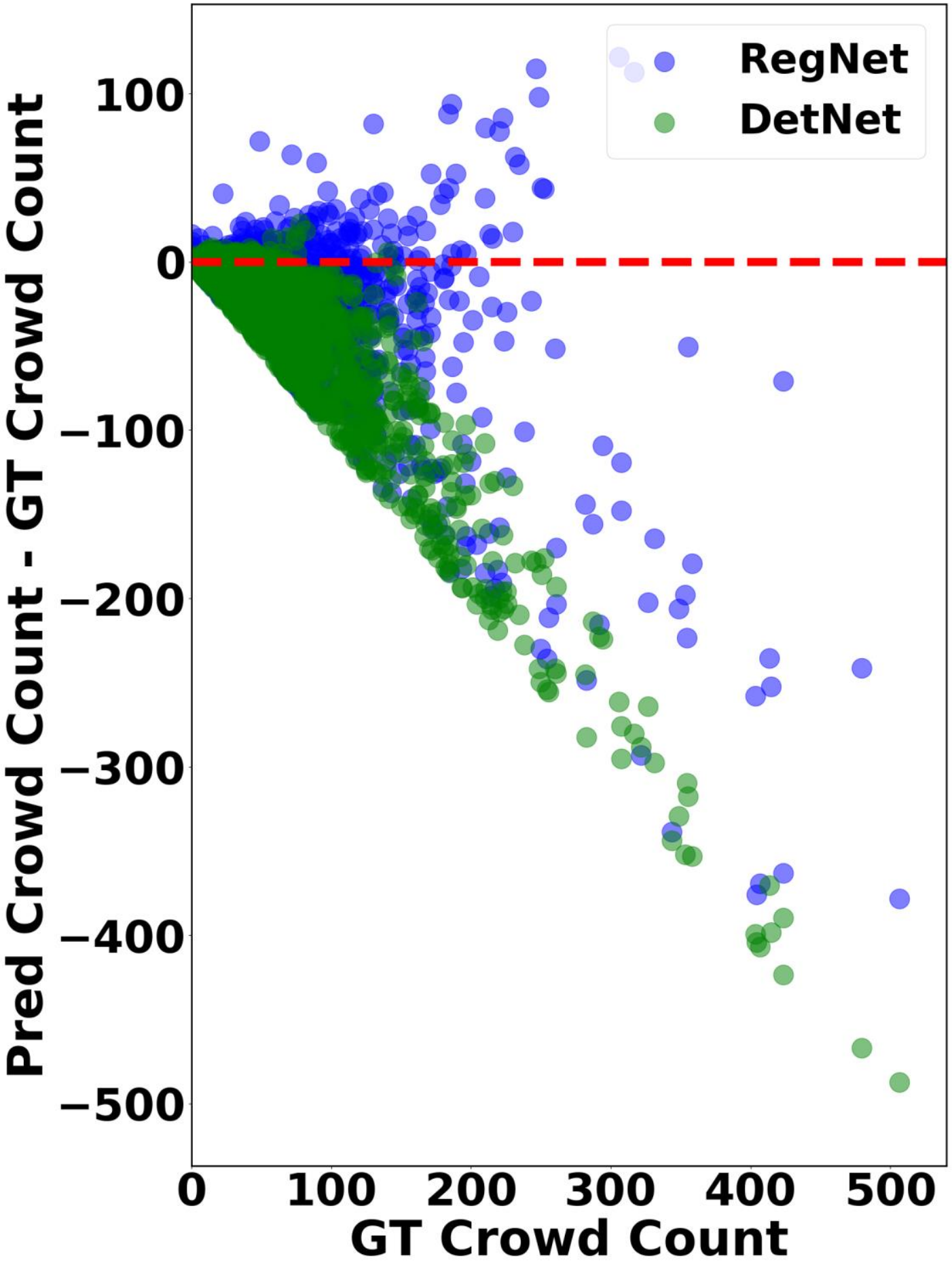} \\
			(a) Visualization Results & (b) Error Distribution \\
    \end{tabular}
\captionof{figure}{(a) Visualization results of an unseen image predicted by a regression-based (top) and a detection-based (bottom) counting model respectively. In the image, `GT' indicates the ground-truth crowd counts, and `Pred' indicates the predicted crowd counts. (b) The error distribution between the predicted and ground truth counts in image patches. [Best viewed in color]}
\label{fig:intro_motivation}
\end{figure}

Crowd counting is to estimate the number of people in images/videos. It is an essential computer vision task widely employed in a batch of valuable multimedia applications, \ie safety management~\cite{boominathan2016crowdnet,huang2019dot}, human behavior modeling, and smart city~\cite{zhang2018wacv,wang2018incremental,tan2019crowd,guo2019dadnet,liu2019point}. Recent advances~\cite{cheng2019learning,zhang2019relational,cheng2019improving,tan2019crowd,liu2019crowd,shi2019revisiting} have witnessed significant progress in the \emph{close-set} counting problem, where crowd counters are normally trained with extensive annotations on person heads in images. This however faces problems when it comes to the \emph{open-set} scenario as it would be very time-consuming and even undesirable to manually annotate hundreds and thousands of new crowd images. On the other hand, directly applying the crowd counters trained on existing data (source) to new data (target) also suffers significant performance degradation owing to the domain shift.

In this paper, we study towards the problem of unsupervised crowd counting in a transfer learning setting where the crowd counter is transferred from a labeled source domain to an unlabeled target domain. Few methods investigated this direction~\cite{wang2019learning,han2020focus}. As a representative, Wang \etal~\cite{wang2019learning} took advantage of a large-scale synthetic dataset and learned to adapt to real crowd counting datasets. Other methods, \ie ~\cite{xu2019iccv,hossainone2019bmvc}, although designed in a single domain setting, have also shown some generalization ability in another domain. These methods are mostly realized via the density regression-based models, where a density distribution is learnt for each crowd image whose integral over the density map gives the total count of that image. A counterpart direction, which has not been investigated in the cross-domain setting, is the detection-based methods~\cite{liu2019point,lian2019density}, where every individual is to be localized in the crowd images. The regression-based methods perform very well in high-density and congested crowd scenes as they do not predict individual locations in particular with occlusions and overlapping. The detection-based methods, on the other hand, are believed to perform better in low-density and sparse crowds with individual locations (see the difference in Fig.~\ref{fig:intro_motivation} (a)). We conducted a further comparison between the regression-based and detection-based methods in the transfer learning setting in Fig.~\ref{fig:intro_motivation} (b): two state-of-the-art methods~\cite{liu2019crowd,liu2019high} were trained on ShanghaiTech SHB~\cite{zhang2016single} {and WiderFace~\cite{yang2016wider}} with full annotations for density regression and face detection, respectively. We directly adopted the trained models to predict on ShanghaiTech SHA and drew the error distribution between the predicted counts and ground truth. The figure shows that two error distributions (denoted by RegNet and DetNet) are clearly separated; DetNet performs better (small errors) than RegNet in low-density areas while its performance significantly drops and underestimates the crowd count in high-density area (green points in Fig.~\ref{fig:intro_motivation}); in contrast, RegNet performs much better than DetNet in relatively high-density area (blue points in Fig.~\ref{fig:intro_motivation}).

Given an unseen crowd image, the observation above tells us that predictions from pre-trained regression- and detection-based counting models provide complementary effect, if we could combine their respective strength in high- and low-density areas. The combination can be realized via the mutual transformations between the predictions of the two models.
Transforming the detection results to combine with the regression is rather straightforward: each individual location can be convolved with a Gaussian kernel to convert to the density distribution~\cite{zhang2016single}. The density estimation result can be enhanced in this way especially in the low-density area~\cite{liu2018decidenet}. On the other hand, transforming the density regression result to combine with the detection, though intuitive, is not explored before.
Analog to convolution and deconvolution, we show that the latter transformation could be formulated as the inverse operation of the former, where we offer a solution in the deep-fashion (see Sec.~\ref{sec:relationmodel}). According to our formulation, the regression-detection mutual transformations are only dependent on the Gaussian kernel used when convolving at each person's location. As long as the Gaussian kernels are adopted in the same rule, the transformation between the regression and detection results can be regarded as two scene-agnostic transformers in crowd counting. They can be learnt in the source and used in the target. 

Given the regression- and detection-based models and their mutual transformers learnt in the source, we propose an unsupervised crowd counting framework in the target via the regression-detection {bi-knowledge} transfer: the two pre-trained regression- and detection-based models in the source are first deployed in the target to obtain initial predictions; the regression and detection predictions pass through their mutual transformers to obtain their counterparts, respectively. The initial and transformed regression (detection) predictions are fused to create the pseudo ground truth for density regression (individual detection). We use the two sets of pseudo ground truth to further fine-tune the regression and detection model in a self-supervised way, respectively. The whole process is repeated for several cycles until the convergence of the two models.

To our knowledge, we are the first to explore the mutual knowledge transfer between regression- and detection-based models towards unsupervised crowd counting. This dual source knowledge is indeed \emph{heterogeneous} as they capture different modalities of the crowd distribution. The contribution of this paper is thus to effectively extract and make use of the \emph{heterogeneous} information from regression- and detection-based models. More specifically,   
\begin{itemize}
\item We introduce a novel unsupervised crowd counting framework to count and localize crowds in the unlabeled target set via the regression-detection bi-knowledge transfer from a labeled source set.  

\item We investigate the mutual transformations between density regression and individual detection, and formulate them as two scene-agnostic transformers in crowd counting. 

\item Thanks to the models and transformers learnt in the source, we propose a self-supervised learning scheme in the target to co-train the regression and detection models with fused pseudo labels and boost the performance of both.
\end{itemize}

We conduct our approach across several standard crowd counting benchmarks, \ie ShanghaiTech~\cite{zhang2016single}, UCF\_CC\_50~\cite{idrees2013multi}, and UCF\_QNRF~\cite{idrees2018composition}, and demonstrate a big margin of improvement over other state-of-the-arts. 
\section{Related Works}
We survey regression- and detection-based crowd counting, as well as cross-domain crowd counting mainly in the deep learning context. 
\subsection{Regression-based crowd counting}
Modern regression-based methods~\cite{zhang2016single,boominathan2016crowdnet,sam2017switching,sindagi2017generating,shi2018crowd,shi2019revisiting,zhao2020eccv} encode the spatial distribution of the crowd into a density map by convolving annotated head points with Gaussian kernels. They learn a mapping from the crowd image to the density map.  The integral of the density map gives the crowd count in the image~\cite{lempitsky2010learning}. Researches in recent trends focused on designing more powerful DNN structures and exploiting more effective learning paradigms~\cite{li2018csrnet,guo2019dadnet,sindagi2019multi,liu2019crowd,cheng2019learning,shi2019revisiting,wan2019adaptive}.
For instance, Guo \etal~\cite{guo2019dadnet} designed multi-rate dilated convolutions to capture rich spatial context at different scales of density maps; Liu \etal~\cite{liu2019crowd} introduced an improved dilated multi-scale structure similarity (DMS-SSIM) loss to learn density maps with local consistency; Xu~\etal~\cite{xu2019iccv} effectively learned to scale multiple dense regions to multiple similar density levels, making the density estimation on dense regions more robust.

Regression-based methods have made remarkable progress when counting the number of persons in crowds. Nevertheless, their performance on low-density crowds are not satisfying~\cite{liu2018decidenet} and they are not capable of providing individual locations in the crowds, which, on the other hand, are believed to be the merits of detection-based crowd counting methods, as specified below. 

\begin{figure*}[t]
    \centering
    \includegraphics[width=\textwidth]{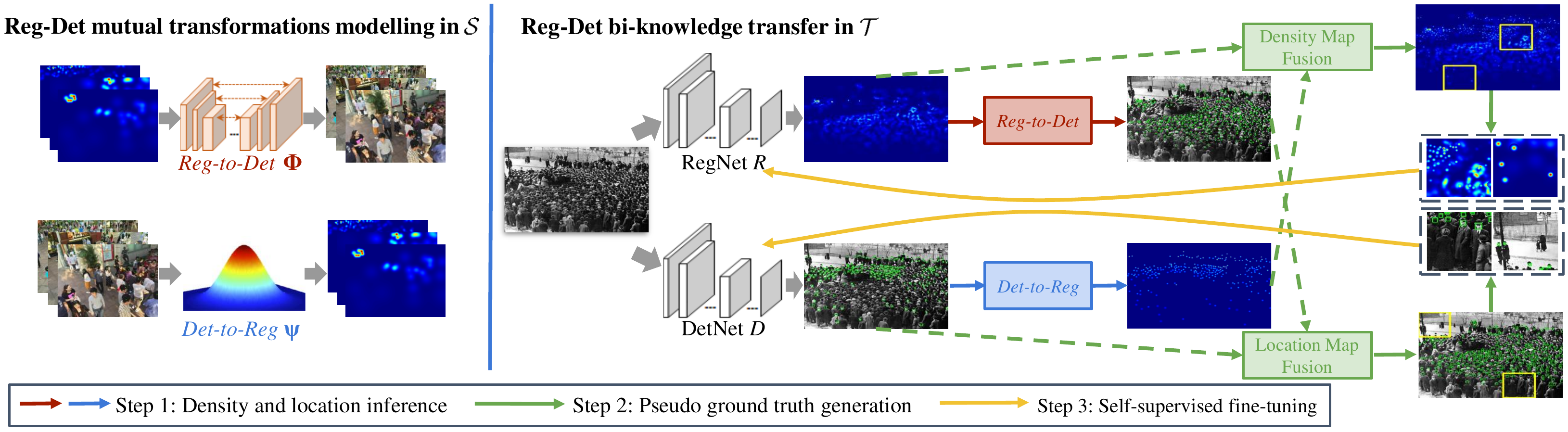}
 \caption{The architecture of our method. It includes two parts: the regression-detection mutual transformations (Left) and the regression-detection bi-knowledge transfer (Right). In the Left, the \textit{Det-to-Reg} transformer is realized via Gaussian convolving at each detected head point; while the \textit{Reg-to-Det} transformer is learnt by an encoder-decoder based model in the source dataset ($\mathcal S$). In the right, given the pre-trained regression and detection models in the source, together with the learnt transformers, it transfers the regression-detection bi-knowledge to an unlabelled target set ($\mathcal T$). It is implemented via an iterative self-supervised way, where, in each cycle, the regression and detection predictions from the pre-trained models are transformed to their counterparts. We fine-tine the two models with their fused pseudo ground truth.}
  \label{fig:framework}
\end{figure*}

\subsection{Detection-based crowd counting.}
Detection-based methods detect precise locations of persons and estimate their counts via the number of detections. They were commonly adopted in relatively low-density crowds~\cite{leibe2005pedestrian,li2008estimating} as the performance would decay severely in high-density crowds with small and occluded persons. 
A recent resurgence of detection-based methods in crowd counting~\cite{stewart2016end,liu2018decidenet,liu2019point,lian2019density} is owing to the advances of object detection in the deep learning context~\cite{redmon2016you,liu2016ssd,ren2015faster,lin2017focal}.   
\cite{liu2018decidenet} trained an end-to-end people detector for crowded scenes depending on annotations of bounding boxes of persons. Liu~\etal~\cite{liu2019point} further designed a weakly supervised detection framework by detecting persons only with point annotations. Besides, Idrees~\etal~\cite{idrees2018composition} introduced a new composition loss to regress both the density and localization maps to infer person location and crowd counts simultaneously. 

Despite the resurgence of detection-based methods, in terms of counting accuracy in dense crowds, they are still not as competitive as those regression-based methods, and often need to be integrated into the latter~\cite{liu2018decidenet,liu2019point,lian2019density}. The integration can be through the attention module in an implicit way~\cite{liu2018decidenet,lian2019density}, while in this paper, we distill the bi-directional knowledge amid regression and detection-based models and learn to explicitly transform the output from one to the other in unsupervised crowd counting.

\subsection{Cross-domain crowd counting}
Most existing crowd counting methods are independently evaluated in single domain dataset with similar crowd scenes~\cite{sam2017switching,li2018csrnet,liu2018decidenet,liu2019point,sindagi2019multi}. Research on cross-domain crowd counting is not largely explored.  
To improve the model generalizability on the new dataset (target) with unseen crowd scenes (unlabeled), Shi \etal~\cite{shi2018crowd} proposed to learn a set of diverse and decorrelated regressors to prevent overfitting in the source dataset;
Xu \etal~\cite{xu2019iccv} addressed the density pattern shift by introducing a learn to scale module (L2SM) to maintain a good transferability across datasets. The crowd information in the target dataset, though unlabeled, is not exploited in~\cite{shi2018crowd,xu2019iccv}, which limits the performance of cross-domain crowd counting. 

In order to make use of data from both source and target, Wang \etal~\cite{wang2019learning} established a large scale synthetic dataset as the source and adopted an unsupervised domain adaptation to reduce the gap between the synthetic source and the real-world target using the Cycle GAN~\cite{zhu2017unpaired}; Han \etal~\cite{han2020focus} instead chose to align domain features in the semantic space using adversarial learning. \cite{wang2019learning,han2020focus} employed density regression models for domain adaption where the target data is taken a whole with a domain label. Unlike them, we propose to model the mutual transformations between the output of density regression and individual detection models in the source, and propagate the regression-detection bi-knowledge through modeled transformers to the target via a self-supervised learning scheme. The transferring process in our work is iterated until the convergence of the performance in the target. 

\section{Method}

\subsection{Problem, motivation and {architecture}}
\label{sec:overview}
Suppose we have a labeled source crowd counting dataset $\mathcal{S}$ and an unlabelled target crowd counting dataset $\mathcal{T}$. Our task is to transfer the knowledge learnt from $\mathcal{S}$ to accurately count and localize persons in $\mathcal{T}$.  

There are two representative manners in conduction of crowd counting: the regression-based and the detection-based. When transferring counting models trained in the source to the target, it is observed (in Fig.~\ref{fig:intro_motivation}) that the regression model performs better in high-density area while the detection model is better in low-density area, which demonstrates a complementary effect from one to the other. In order to combine their strength, this dual source knowledge needs to be transformable between each other, and transferable from the source to the target. Transforming the detection result to the regression result is rather a standard procedure: using a Gaussian kernel to convolve at each detected individual location~\cite{zhang2016single,liu2018decidenet}. Its inverse problem, transforming the regression result to the detection result, however, has not been exploited before. We show in Sec.~\ref{sec:relationmodel} a modern solution, analog to deconvolution in deep learning, by modelling it in deep neural networks. The motivation of this paper is thus to take advantage of the knowledge of regression and detection models learnt in the source and mutual transfer and distill them in the target to enhance unsupervised crowd counting.

Fig.~\ref{fig:framework} presents an overview of our method. It consists of two parts: the detection-regression mutual transformation modeling in the source and the iterative
self-supervised learning in the target. The former part models mutual transformations, \ie the \textit{Det-to-Reg} and  \textit{Reg-to-Det} transformers in the source. {The detection- and regression models are also trained in the source.} The latter part conducts bi-knowledge transfer of detection and regression models in the target with an iterative self-supervised learning scheme.

\subsection{Base networks}
\label{sec:basenetworks}
Before starting the technical details, we first introduce the regression and detection networks employed in this paper. They are learnt with ground truth annotations in the source dataset. 

\para{Regression network~} We choose the deep structured scale integration network (DSSINet)~\cite{liu2019crowd} as our regression network $R$. It is good at addressing the scale shift problem by using conditional random fields (CRFs) for messaging passing among multi-scale features. DSSINet takes VGG16 ~\cite{simonyan2014very} as its backbone and is trained on $\mathcal{S}$ with the proposed dilated multi-scale structural similarity loss. The output of the network is a crowd density map $M^R$.

\para{Detection network~}
We adopt the center and scale prediction (CSP) based pedestrian detector~\cite{liu2019high} as our detection network ${D}$. CSP is an anchor-free keypoint based detector that predicts the center point and scale of each pedestrian. CSP takes ResNet-50~\cite{he2016deep} as its backbone and is trained with the cross-entropy classification loss and the smooth-L1 loss. 
The output of the network consists of an individual localization map $M^D$ (0-1 map) and a scale map $M^S$ indicating the person's location and size information, respectively.

\subsection{Regression-detection mutual transformations}
\label{sec:relationmodel}
In this section, we formulate the mutual transformations between the output of the regression and detection models as \emph{Det-to-Reg $\Psi$} and \emph{Reg-to-Det $\Phi$}, respectively. 

\heading{Det-to-Reg $\Psi$} refers to the transformation from the individual localization and scale maps (${{M}^{D}}$, ${{M}^{S}}$) to the crowd density map ${M}^{R}$. This can be achieved in a rather standard way following~\cite{liu2018decidenet,zhang2016single}: we convolve at each nonzero point of the individual localization map ${M}^{D}$ with a Gaussian kernel $G_{\sigma _{j}}$ to produce the crowd density map ${M^R}(z)$,
\begin{equation}
\label{equ:Det-to-Reg}
{M^R}(z) = \Psi (M^D) = \sum_{j=1}^{H} \delta(z-z_j)*G_{\sigma _{j}}(z), \end{equation}
where $z_j$ signifies the $j$-th nonzero pixel in $M^D$ and $H$ is the total number of nonzero pixels (heads) in $M^D$. $\sigma _{j}$ is proportional to the person scale value at point $j$, \ie $\sigma _{j} \propto M^S_j$. 
In practice, many crowd counting datasets only have head point annotations ($M^D$) available. When generating the ground truth density map, $\sigma_j$  is either fixed (in sparse crowd scenes) or approximated via the distance ${d}_j$ from person $j$ to his/her nearest neighbors (in dense crowds)~\cite{zhang2016single}, \ie $\sigma _{j} = \beta {d}_j$. 

\heading{Reg-to-Det $\Phi$} models the transformation from the crowd density map ${{M}^{R}}$ to the individual localization and scale maps (${{M}^{D}}$, ${{M}^{S}}$). Based on the similar spirit above, ${{M}^{S}}$ can be easily estimated by referring to the distance from each person detected in ${{M}^{D}}$ to its nearest neighbors, or if we have another $M^S$, \eg detection adapted from source to the target (see Sec.~\ref{sec:transfer}), the transformed scale map from $M^R$ can be simply referred to values in the adapted scale map. Thus, the real target is to find the projection $\Phi$ from $M^R$ to ${{M}^{D}}$, ${{M}^{D}} = \Phi({M}^{R})$. This is an inverse operation to Eq.~\ref{equ:Det-to-Reg}. Recovering $M^D$ is indeed to find $\Phi$ to minimize the equation, 
\begin{equation}
\label{eq:lassominimiazation}
{\Phi}^* = \min\limits_{\Phi} \| M^D - \Phi(M^R) \|_2
\end{equation}

Without loss of generality, the \textit{Det-to-Reg} transformer $\Psi$ in Eq.~\ref{equ:Det-to-Reg} can be approximated as matrix multiplication analog to the CUDA implementation of convolution in deep learning. In light of the learnable convolution and deconvolution, we can hence model $\Phi$ in Eq.~\ref{eq:lassominimiazation} as a non-linear mapping of the inverse operation of $\Psi$ in the deep learning context.
 
We employ the nested UNet~\cite{zhou2019unet++} as an encoder-decoder with dilated VGG-16 structure~\cite{li2018csrnet} to learn the mapping from ${M}^{R}$ to ${{M}}^{D}$. The output of the encoder-decoder $\Phi({{M}}^{R})$ is enforced to be as close as ${{M}}^{D}$ with an MSE loss applied to every image:
\begin{equation}
\label{equ:mse}
L_{MSE} = \| M^D - \Phi(M^R) \|_2 = 
\frac{1}{N} \sum _{i=1}^{N} ({M}^{D}_{i} - \Phi(M^R_i))^2
\end{equation}
where $i$ signifies the $i$-th pixel in the map, and there are in total $N$ pixels in the image.  

Although a crowd in an image can be very dense, the localization of each individual is marked with only one pixel in $M^D$, meaning that $M^D$ is rather sparse, $H \ll N$. To balance the loss contributions between non-zero and zero pixels in $M^D$, we are motivated to adapt the focal loss~\cite{lin2017focal}, which specifically copes with the unbalance issue in object detection, into a focal MSE Loss in our scenario: 
\begin{equation}
\label{equ:focal}
L_{Focal-MSE} = \frac{1}{N} \sum _{i=1}^{N} 
  \alpha_i(1-p_i)^{\gamma} ({M^D_i} - \Phi(M^R_i))^2,
\end{equation}
where 
\begin{equation}
\label{equ:pt}
p_i =\left\{
             \begin{array}{ll}
             {\rm sigmoid} (\Phi({M^R_i})) 
                 &  {\rm if}\quad {{M}}^{D}_{i} ==1  \\
             1 - {\rm sigmoid}(\Phi({M^R_i}))         &  {\rm otherwise}
             \end{array}
\right.
\end{equation}
and
\begin{equation}
\label{equ:alpha}
\alpha_i =\left\{
             \begin{array}{ll}
             1
                 &  {\rm if}\quad M^D_i ==1  \\
             0.1         &  {\rm otherwise}
             \end{array}
\right.
\end{equation}
$\alpha_i$ is a weighting factor that gives more weights on the nonzero pixels of $M^D$ as its number is much less than that of the zero pixels. $(1-p_i)^{\gamma}$ is a modulating factor that reduces the loss contribution from easy pixels (\eg $\Phi(M^R_i)$ with very large value at $M^D_i == 1$ or very small value at $M^D_i == 0$) while focuses on those hard pixels. $\gamma$ is a parameter ($\gamma = 2$ in practice) to smoothly adjust the rate for easy pixels to be down-weighted.

We also adopt the Dilated multiscale structural similarity (DMS-SSIM) loss in~\cite{liu2019crowd} $L_{DMS-SSIM}$ to enforce the local patterns (mean, variance and covariance) of $\Phi(M^R)$ visually similar to $M^D$. Its parameter setting is the same to \cite{liu2019crowd}. To this end, the final objective function for the \textit{Reg-to-Det} transformer $\Phi$ is
\begin{equation}
\label{equ:Reg-to-Det_loss}
L_{\Phi} = L_{Focal-MSE} +  L_{DMS-SSIM}.
\end{equation}

The encoder-decoder $\Phi$ is trained in the source $\mathcal S$ with ground truth individual localization map and density map. The output matrix is binarized with a threshold of 0.2. We further merge redundant non-zero pixels within the local area ($10\times 10$) to obtain the final binary matrix. 

Both $\Psi$ and $\Phi$ are scene-agnostic transformations in crowd counting. As long as the Gaussian kernel is designed with the same rule, we can use $\Psi$ and $\Phi$ to distill the knowledge between the regression and detection models in the target set $\mathcal T$.   

\subsection{Regression-detection bi-knowledge transfer}
\label{sec:transfer}
In this session, we transfer the knowledge learnt from the labeled source dataset to the unlabeled target dataset. The transfer is bi-directional between the regression and detection models. It is realized in an iterative \emph{self-supervised} way initiated from the pre-trained regression and detection models, $R_0$ and $D_0$, in the source (see Sec.~\ref{sec:basenetworks}). Without loss of generality, we use $R_t$ and $D_t$ to denote the regression and detection model in the $t$-th cycle. Three steps are carried out in each \emph{self-supervised learning} cycle: \textbf{1)} $R_t$ and $D_t$ are used to infer the crowd density and location maps for each image $I$ in $\mathcal{T}$, respectively; \textbf{2)} the inferred maps from $D_t$ ($R_t$) is fused with the transformed maps using $\Psi$($\cdot$) ($\Phi$($\cdot$)) to generate pseudo ground truth; \textbf{3)} $R_t$ and $D_t$ are fine-tuned with the regression and and detection pseudo ground truth in $\mathcal{T}$ to further update to $R_{t+1}$ and $D_{t+1}$. The whole process iterates for several rounds until the convergence of $R$ and $D$ in $\mathcal{T}$. Below we detail the three steps. 

\heading{Crowd density and location inference.} Given $R_t$ and $D_t$, we use $R_t(I)$ and $D_t(I)$ to indicate the crowd density and individual localization maps per image $I$. They can be easily obtained by forwarding the network with $R_t$ and $D_t$ once, respectively. 

\heading{Pseudo ground truth generation.} Referring to Fig.~\ref{fig:intro_motivation}, the dual  prediction $R_t(I)$ and $D_t(I)$ on image $I$ complements each other. To take advantage of both, we propose to distill the knowledge from one to the other to further fine-tune the model, respectively. Using the mutual transformations discussed in Sec.~\ref{sec:relationmodel}, we can obtain the counterpart of $R_t(I)$ and $D_t(I)$ via $\Psi(D_t(I))$ and $\Phi(R_t(I))$. $R_t(I)$ is then fused with $\Psi(D_t(I))$, $D_t(I)$ is with $\Phi(R_t(I))$, correspondingly. 

Regarding the fusion between $R_t(I)$ and $\Psi(D_t(I))$, we propose to use the detection confidence weight map $W_t$ to act as a guidance. $W_t$ is modified such that within each $k \times k$ (same $k$ for the Gaussian kernel) area of a detection center, the weights are set the same to the center weight. The fused regression result $M^{R_t}$ is given by,
\begin{equation}
\label{equ:pseudo_denmap}
M^{R_t} = (1-{W}_t) \cdot {R}_t({I})  +  {W}_t \cdot  \Psi({D}_t({I})).
\end{equation}
The reason to use $W_t$ is that the detector $D_t$ normally performs better in low-density area with high confidence scores; thus its transformed regression result $\Psi({D}_t({I}))$ contributes more in the low-density area of $M^{R_t}$ if multiplying by $W_t$; the original ${R}_t({I})$ instead contributes more in the high-density area in Eq.~\ref{equ:pseudo_denmap}.  

Regarding the fusion between $D_t(I)$ with $\Phi(R_t(I))$, the spirit is similar: 
the transformed detection result $\Phi(R_t(I))$ from $R_t(I)$ produces more detection than $D_t(I)$ in the high-density area and while $D_t(I)$ normally produces less compared to the ground truth (see Fig.~\ref{fig:intro_motivation} (b)). We can simply fuse the detection from $D_t(I)$ with $\Phi(R_t(I))$ followed up by non-maximum suppression (NMS)~\cite{neubeck2006efficient}, which should result in adequate detection in both high- and low-density area. We denote by $M^{D_t}$ the final detection result. Notice that $\Phi(R_t(I))$ only produces individual center locations but not scales (sizes). {In order to restore complete bounding boxes, we find the corresponding scales $M^{S_t}$ using the original scale map $M^S$ from $D_t$
in the lower half of the image and nearest neighbor distances in the upper half of the image}.

Having received ${M}^{R_t}$ and ${M}^{D_t}$ ($M^{S_t}$), 
we select two patches of size $224 \times 224$ from each image, their pseudo labels are cropped correspondingly from the map. For $M^{R_t}$, we traverse all the non-overlapped patches with their densities and select the ones with medium (high)-density area as the prediction on the extremely high-density or low-density area would not be accurate. {Concretely, this is achieved by randomly selecting from the density ranges 50\% - 70\%}. For  $M^{D_t}$, we also traverse all non-overlapped patches and find ones with average detection confidence scores being the highest.   

\heading{Self-supervised fine-tuning.} We fine-tune $R_t$ and $D_t$ with the samples selected from every image alongside their pseudo ground truth obtained above. The updated models are denoted by $R_{t+1}$ and $D_{t+1}$. Having the updated model, we could repeat the whole process to re-select samples from images and re-train the two models until their convergence at the $T$-th iteration, $R_T$ and $D_T$. 

At the testing stage, for each image we predict its results using both $R_T$ and $D_T$. We get the crowd counting and detection results by merging the outputs of $R_T$ and $D_T$ with the same procedure as in training. 

\section{Experiments}

\tabcolsep=2pt
\begin{table*}[t] 
\caption{Comparisons with the state-of-the-art methods in the transfer setting. \bc{$\mathtt{Syn}$} denotes transfer learning using a large size synthetic dataset. \rc{$\mathtt{Real}$} denotes transfer learning using a real source dataset. The methods with $*$ indicate the source is the synthetic dataset GCC.}
\centering
\label{tab:sota}
\resizebox{1\linewidth}{!}{
\begin{tabular}{l |c |ccc| ccc| ccc| ccc |ccc }
\toprule
\multirow{2}{*}{\textbf{Method}} & 
&\multicolumn{3}{c|}{\textbf{A$\rightarrow$B}}  
&\multicolumn{3}{c|}{\textbf{A$\rightarrow$C}} 
&\multicolumn{3}{c|}{\textbf{A$\rightarrow$Q}} 
&\multicolumn{3}{c|}{\textbf{B$\rightarrow$A}} 
&\multicolumn{3}{c}{\textbf{B$\rightarrow$Q}}\\
    & &  MAE$\downarrow$ & MSE$\downarrow$ & mAP$\uparrow$ & MAE$\downarrow$ & MSE$\downarrow$ & mAP$\uparrow$ & MAE$\downarrow$ & MSE$\downarrow$ & mAP$\uparrow$ &  MAE$\downarrow$ & MSE$\downarrow$ & mAP$\uparrow$ & MAE$\downarrow$ & MSE$\downarrow$ & mAP$\uparrow$\\
            \midrule
            \midrule

Cycle GAN$^{*}$~\cite{zhu2017unpaired} & \bc{$\mathtt{Syn}$} 
&25.4   &39.7   &--  
&404.6  &548.2  &-- 
&257.3  &400.6  &-- 
&143.3  &204.3  &-- 
&257.3  &400.6  &-- \\

SE Cycle GAN$^{*}$~\cite{wang2019learning} & \bc{$\mathtt{Syn}$} 
&19.9   &28.3   &--  
&373.4  &528.8  &-- 
&230.4  &384.5  &-- 
&123.4  &193.4  &-- 
&230.4  &384.5  &-- \\

SE+FD$^{*}$~\cite{han2020focus} & \bc{$\mathtt{Syn}$} 
&16.9   &24.7   &--  
&--     &--     &-- 
&221.2  &390.2  &-- 
&129.3  &187.6  &-- 
&221.2  &390.2  &-- \\

\midrule

MCNN~\cite{zhang2016single} & \rc{$\mathtt{Real}$}
&85.2   &142.3  &--  
&397.7  &624.1  &-- 
&--     &--     &-- 
&221.4  &357.8  &-- 
&--     &--     &-- \\

D-ConvNet-v1~\cite{shi2018crowd} & \rc{$\mathtt{Real}$}
&49.1   &99.2   &--  
&364    &545.8  &-- 
&--     &--     &-- 
&140.4  &226.1  &-- 
&--     &--     &-- \\

L2R~\cite{liu2018leveraging} & \rc{$\mathtt{Real}$}
&--     &--     &--  
&337.6  &434.3  &-- 
&--     &--     &-- 
&--     &--     &-- 
&--     &--     &-- \\

SPN+L2SM~\cite{xu2019iccv}  & \rc{$\mathtt{Real}$}
&21.2   &38.7   &--  
&332.4  &425.0  &-- 
&227.2  &405.2  &-- 
&126.8  &203.9  &-- 
&--     &--     &-- \\

\midrule

RegNet~\cite{liu2019crowd} & \rc{$\mathtt{Real}$}
&21.65  &37.56  &-- 
&419.53 &588.90 &--
&198.71 &329.40 &-- 
&148.94 &273.86 &--  
&267.26 &477.61 &-- \\

DetNet~\cite{liu2019high}  & \rc{$\mathtt{Real}$}
&55.49  &90.03  &0.571
&703.72 &941.43 &0.258
&411.72 &731.37 &0.404
&242.76 &400.89 &0.489
&411.72 &731.37 &0.404\\

\textbf{Ours}  & \rc{$\mathtt{Real}$}
&\textbf{13.38}  &29.25  & \textbf{0.757}   
&368.01&518.92&\textbf{0.518}
&\textbf{175.02}&\textbf{294.76}&\textbf{0.546}
&\textbf{112.24} &218.18 & \textbf{0.661} 
&\textbf{211.30} & \textbf{381.92} & \textbf{0.535}\\
\bottomrule
\end{tabular}}
\end{table*}

\subsection{Datasets}
\heading{ShanghaiTech~\cite{zhang2016single}.} It consists of 1,198 annotated images with a total of 330,165 people with head center annotations. This dataset is split into two parts: \textbf{SHA} and \textbf{SHB}. The crowd images are sparser in SHB compared to SHA: the average crowd counts are 123.6 and 501.4, respectively. We use the same protocol as in \cite{zhang2016single} that 300 images are for training and 182 images for testing in SHA; 400 images are for training and 316 images for testing in SHB.

\heading{UCF\_CC\_50~\cite{idrees2013multi}.} It has 50 images with 63,974 head center annotations in total. The headcount ranges between 94 and 4,543 per image. The small dataset size and large variance make it a very challenging counting dataset. Following~\cite{idrees2013multi}, we perform 5-fold cross-validation to report the average test performance.

\heading{UCF\_QNRF~\cite{idrees2018composition}.} It is a large crowd counting dataset with 1535 high-resolution images and 1.25 million head annotations, among which 334 images are used as the testing set. The dataset contains extremely congested scenes where the maximum count of an image can reach 12865.

\subsection{Implementation details and evaluation protocol}
\heading{Implementation details.} Training details of DSSINET for regression and CSP for detection in the source follow the same protocol with~\cite{liu2019crowd} and ~\cite{liu2019high}, as specified in Sec.~\ref{sec:basenetworks}.
Notice that the source crowd counting datasets do not provide bounding box annotations for training CSP, we thus pre-train it on a face detection dataset {WiderFace~\cite{yang2016wider} and adopt it in the source with point annotations following~\cite{liu2019point}}.

To train the encoder-decoder for $\Phi$, we randomly crop 224 $\times$ 224 patches from the ground truth density maps in $\mathcal S$ for training. We initialize the first 10 convolutional layers of encoder-decoder with the weights from a VGG16 net~\cite{simonyan2014very} pre-trained on ILSVRC classification task. The rest convolutional layers are initialized via a Gaussian distribution with zero mean and standard deviation of $1\times 10^{-2}$. The encoder-decoder is optimized by Adam with a learning rate of $1 \times 10^{-5}$. 

For fine-tuning the regression model in the target set $\mathcal T$, Adam optimizer is used with a learning rate $1 \times 10^{-6}$. During fine-tuning, only weights of the last two layers are updated. This ensures that the learnt knowledge in the source is preserved while the last two layers adapt to the new data.

For fine-tuning the detection model in $\mathcal T$, we also use Adam optimizer with a learning rate of $1 \times 10^{-5}$.

\heading{Evaluation protocol} We evaluate both the person localization and counting performance. To measure the counting performance, we adopt the commonly used mean absolute error ({MAE}) and mean square error ({MSE})~\cite{sam2017switching,sindagi2017generating} to compute the difference between the counts of ground truth and estimation.

To report localization performance, we measure the mAP for person head localization. Those predicted head points within a particular distance of $c$ pixels to its nearest ground truth point are regarded as \emph{true positives}. For a certain ground truth point, if there exist duplicate predictions that satisfy the condition, we choose the one with the highest score as true positive. Others are taken as \emph{false positives}. Average precision (AP) is computed for every $c$ and the mAP is obtained as the average value of AP with various $c$. $c$ is varied from 1 to 100 as in~\cite{idrees2018composition}.

\tabcolsep=1pt
\begin{figure*}[tb]
	\centering
		\begin{tabular}{lcccc}
			\rotatebox{90}{  DetNet} &
			\includegraphics[width=0.22\textwidth, height=35mm]{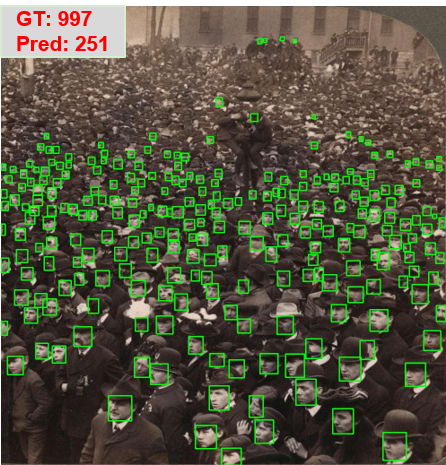} &
			\includegraphics[width=0.22\textwidth, height=35mm]{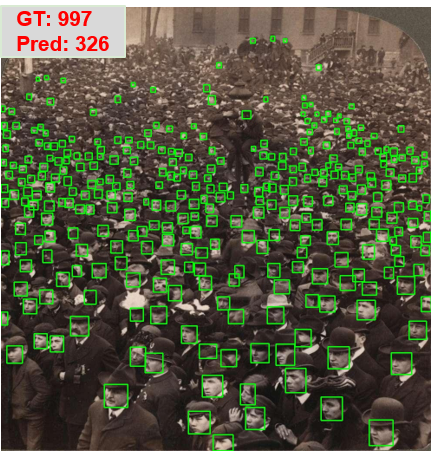} &
			\includegraphics[width=0.22\textwidth, height=35mm]{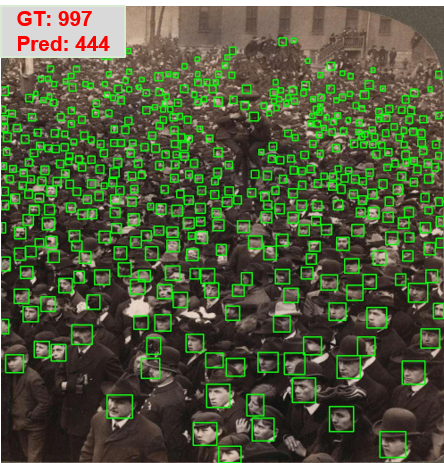} &
			\includegraphics[width=0.22\textwidth, height=35mm]{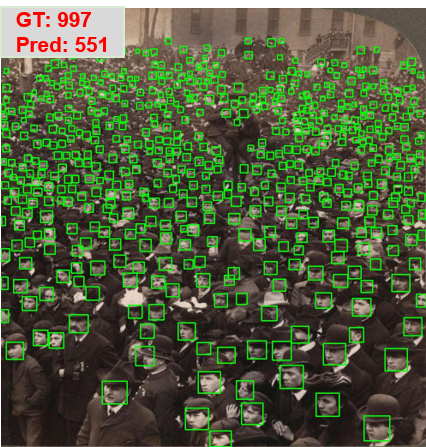} \\
			\rotatebox{90}{  RegNet} &
			\includegraphics[width=0.22\textwidth, height=35mm]{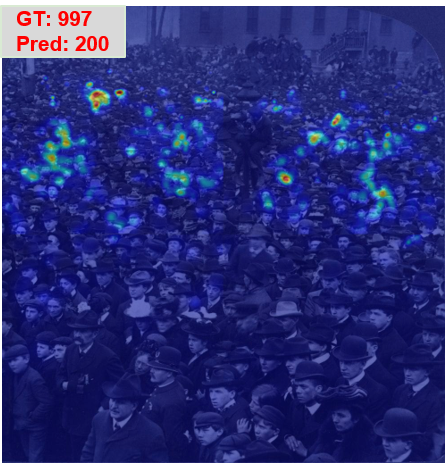} &
			\includegraphics[width=0.22\textwidth, height=35mm]{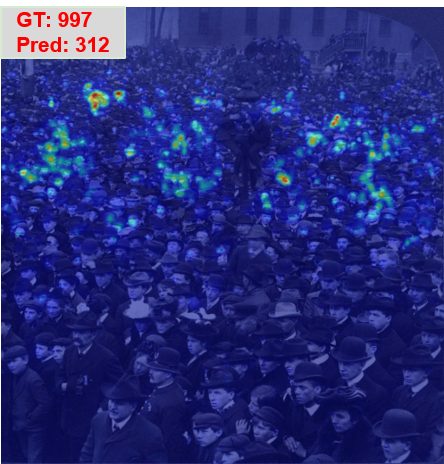} &
			\includegraphics[width=0.22\textwidth, height=35mm]{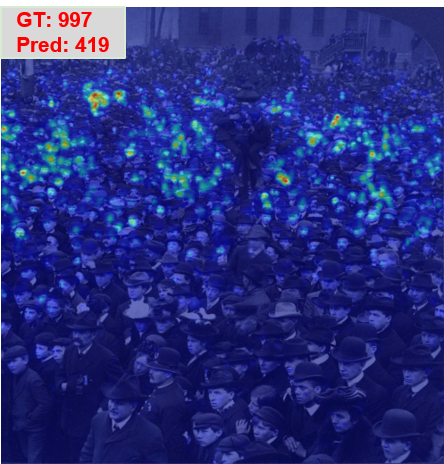} &
			\includegraphics[width=0.22\textwidth, height=35mm]{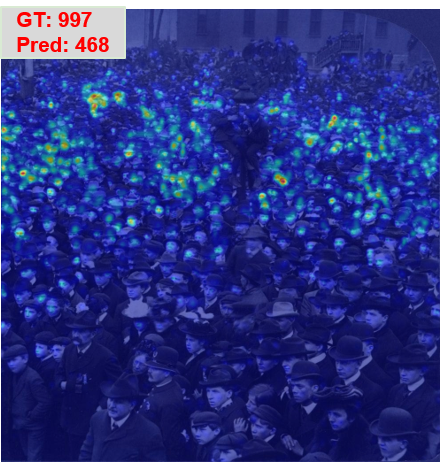} \\
			 & Iteration 0  & Iteration 1  & Iteration 2 & Iteration 3\\
	\end{tabular}
   \caption{An example of visualization results of our method when transferring from SHB to SHA in different iterations.}
\label{fig:existing}
\end{figure*}

\subsection{Comparisons with the State-of-the-arts}
To show the effectiveness of our method, we compare it with other state-of-the-arts in cross-domain setting~\cite{wang2019learning,zhu2017unpaired,han2020focus,zhang2016single,shi2018crowd,liu2018leveraging,xu2019iccv}. Methods including Cycle GAN~\cite{zhu2017unpaired}, SE Cycle GAN~\cite{wang2019learning}, and SE+FD~\cite{han2020focus} transfer the knowledge from a very large-scale synthetic dataset GCC~\cite{wang2019learning}, which contains 15,212 high resolution images. Methods such as MCNN~\cite{zhang2016single}, D-ConvNet-v1~\cite{shi2018crowd}, L2R~\cite{liu2018leveraging}, and SPN+L2SM~\cite{xu2019iccv} learn the model from a real source dataset like ours. We present the results of transferring from SHA (A) to SHB (B), UCF\_CC\_50 (C), and UCF\_QNRF (Q), as well as from SHB to SHA and UCF\_QNRF in Table~\ref{tab:sota}. 

When transferring between SHA and SHB, our method performs the lowest MAE, \ie 13.38 for A $\rightarrow$ B and 112.24 for B $\rightarrow$ A, which improves other SOTA with a big margin. When transferring in a more difficulty setting, \ie from  SHA/SHB to the large-scale dataset UCF\_QNRF, our method produces significantly better results on both A $\rightarrow$ Q and B $\rightarrow$ Q over others; Compared with \cite{zhu2017unpaired,wang2019learning,han2020focus}, which learn from the much larger and more diverse synthetic source GCC, our method shows very satisfying transferability. The result of A $\rightarrow$ C is relatively inferior to the best in prior arts. This is caused by the inferior detection result as crowd scenes in UCF\_CC\_50 are too congested to obtain satisfying localization results.

Overall, our method achieves the best counting accuracy in most of the transfer settings. More importantly, we would like to point out that our method is also capable of providing precise individual localization of the crowds (see mAP in Table~\ref{tab:sota}), which is another advantage over the state of the art.

\subsection{Ablation Study}
The first ablation study is focused on the essential idea of regression-detection {bi-knowledge transfer} for the unsupervised crowd counting. We present the results of using either the regression network~\cite{liu2019crowd} or detection network~\cite{liu2019high} (denoting by RegNet and DetNet, respectively) trained from the source to directly predict the crowd density or individual localization in the target. The results are shown in  Table~\ref{tab:sota}. Compared to our method, they are clearly inferior in terms of both counting and detection accuracy. This also illustrates the effectiveness of our method combining the strength of both models and delivering much more competitive results.      

Next, we study the combination details between the regression and detection models as proposed in Sec.~\ref{sec:transfer}. First, the output of $R_t$ and $D_t$ are not fused via the proposed regression-detection transformers (\eg Eq.\ref{equ:pseudo_denmap}) but instead they are fine-tuned with their own pseudo ground truth. Patch sampling within the pseudo ground truth maps follows the same procedure as in Sec.~\ref{sec:transfer} to choose the reliable and discriminative patches per image. The training iterates several cycles until the convergence of $R_T$ and $D_T$. We present their results separately in Table~\ref{tab:ablation} with the notation ours \emph{w/o fusion}. Comparison is made in the B $\rightarrow$ A setting. For instance for the regression result, the MAE and MSE are 146.78 and 275.35, respectively; by using the fusion, they can be reduced to 112.24 (-34.54) and 218.18 (-57.17). Similarly, for the detection result, the MAE and MSE significantly decrease 127.63 and 184.62 points, respectively; the AP increases 19.4\%. This demonstrates the importance of knowledge distillation between the regression and detection networks.   

Another detail of the regression and detection combination lies in patch sampling strategy, where we propose to sample patches from the medium-density area for the regression fine-tuning; from the high confidence area for the detection fine-tuning. To justify this strategy, we compare it with random sampling in Table~\ref{tab:ablation}. It can be seen that, our results are significantly better than random sampling (denoted by ours \emph{w/ RS}) for both regression and detection results. Notice that the outputs of the regression and detection networks in ours \emph{w/ RS} are fused in the same way as with ours.  

Last, we study the iterations of the proposed self-supervised learning in Sec.~\ref{sec:transfer}. 
Table~\ref{tab:iteration} illustrates the results of the regression and detection along with the increase of iterations. It is clear that both the regression and detection benefit from the iterative co-training as the iteration increases. The performance gets stable after several iterations, when the improvement becomes marginal on both regression and detection. Normally, we stop after 4 iterations.

\tabcolsep=6pt 
\begin{table}[t] 
\caption{Ablation study on regression-detection {bi-}knowledge transfer. \emph{w/o Fusion} means regression and detection outputs are not fused; \emph{w/ RS} means training patches are randomly selected from the fused density or localization maps.} 
\centering
\label{tab:ablation}
\resizebox{1\linewidth}{!}{
\begin{tabular}{lcc|ccc}
\toprule
\multirow{2}{*}{B$\rightarrow$A}    &  \multicolumn{2}{c|}{Regression} &  \multicolumn{3}{c}{Detection} \\
    &  MAE$\downarrow$ & MSE$\downarrow$ &  MAE$\downarrow$ & MSE$\downarrow$ & mAP$\uparrow$ \\
            \midrule
            Ours \emph{w/o Fusion}           &146.78  &275.35 &251.75  &407.04  &0.467 \\
            Ours \emph{w/ RS}       &152.19  &281.05  &165.80 &302.90  & 0.609\\ %
            Ours                      &112.24  &218.18 &124.12  &222.42  &0.661 \\

\bottomrule
\end{tabular}}
\end{table}

\tabcolsep=8pt
\begin{table}[h] 
\caption{Ablation study on self-supervised learning iterations. }
\centering
\label{tab:iteration}
\resizebox{1\linewidth}{!}{
\begin{tabular}{lcc|ccc}
\toprule
  \multirow{2}{*}{B$\rightarrow$A}  &  \multicolumn{2}{c|}{Regression} &  \multicolumn{3}{c}{Detection} \\
    &  MAE$\downarrow$ & MSE$\downarrow$ &  MAE$\downarrow$ & MSE$\downarrow$ & mAP$\uparrow$ \\
            \midrule
            Iteration 0         &148.94  &273.86 &242.76  &400.89  &0.489 \\%
            Iteration 1         &136.81  &258.38 & 172.36 & 305.04 & 0.598\\ 
            Iteration 2         &125.86  &228.31 &138.18  &251.75  &0.621 \\  
            Iteration 3         &114.06  &222.55 &126.86  &231.45  &0.658 \\ 
            Iteration 4         &112.24  &218.18 &124.12  &222.42  &0.661 \\ 
            Iteration 5      & 113.06 & 220.17 &125.12&213.26&0.665\\
\bottomrule
\end{tabular}}
\end{table}

\tabcolsep=1pt
\begin{figure*}[t]
	\centering
		\begin{tabular}{cccc}
			\includegraphics[width=0.22\textwidth, height=31mm]{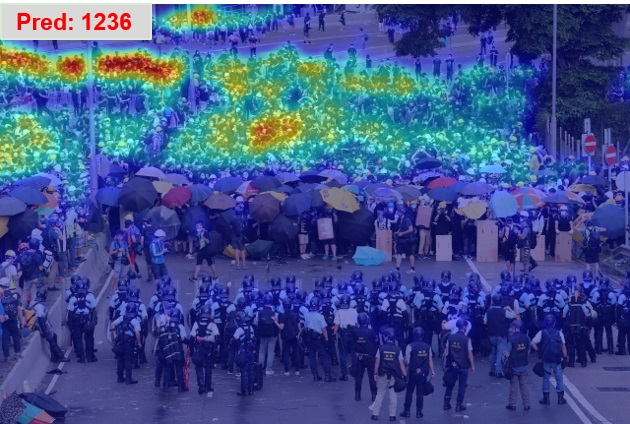} &
			\includegraphics[width=0.22\textwidth, height=31mm]{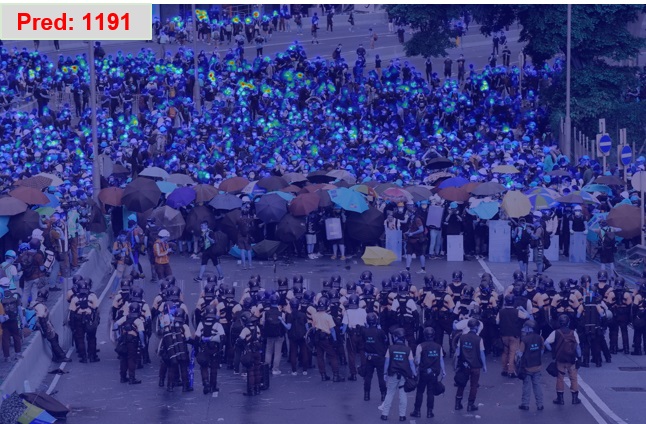} &
			\includegraphics[width=0.22\textwidth, height=31mm]{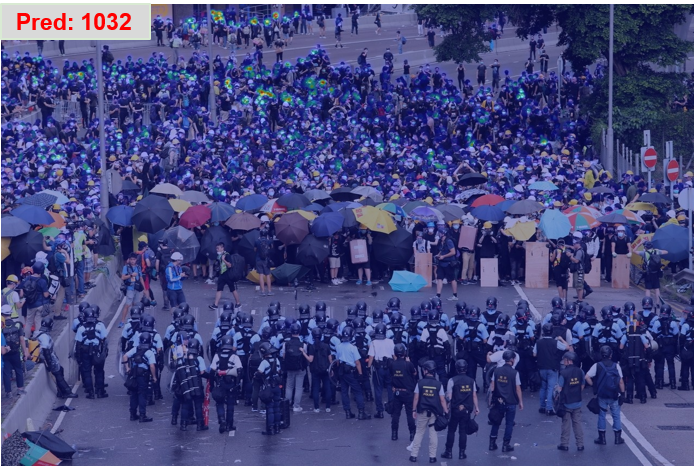} &
			\includegraphics[width=0.22\textwidth, height=31mm]{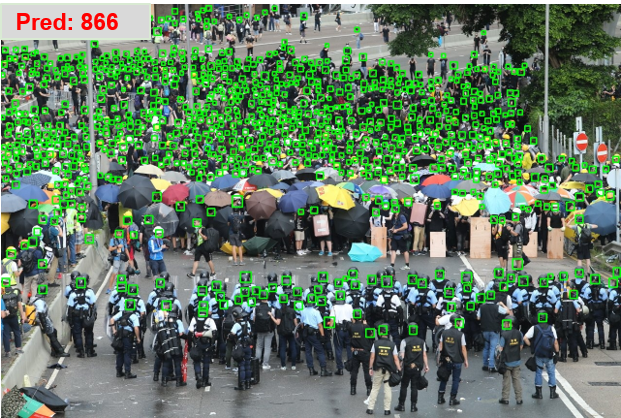} \\
			\includegraphics[width=0.22\textwidth, height=31mm]{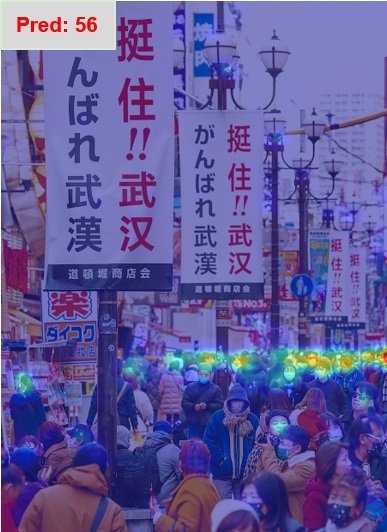} &
			\includegraphics[width=0.22\textwidth, height=31mm]{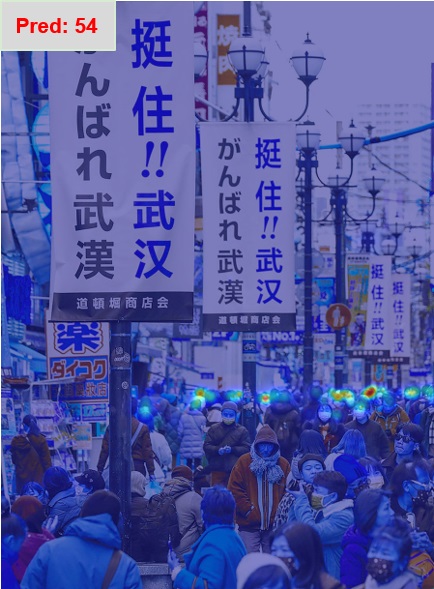} &
			\includegraphics[width=0.22\textwidth, height=31mm]{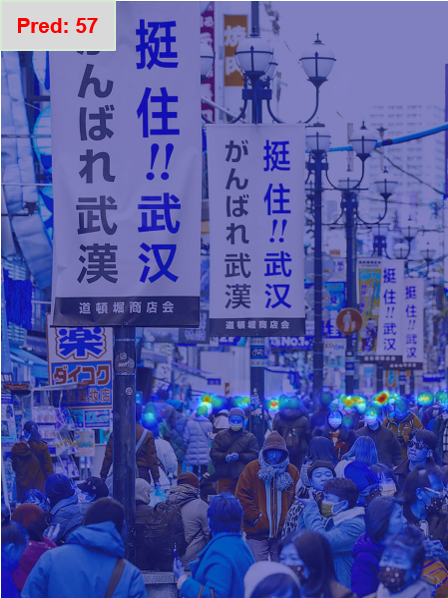} &
			\includegraphics[width=0.22\textwidth, height=31mm]{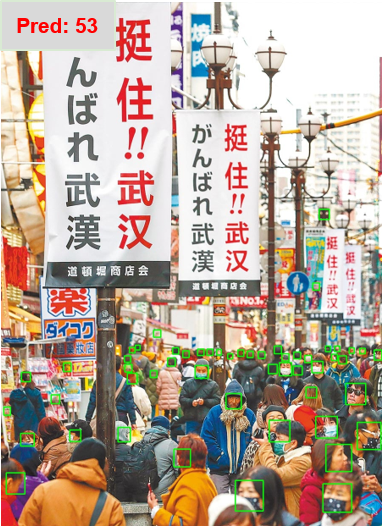} \\
			\includegraphics[width=0.22\textwidth, height=31mm]{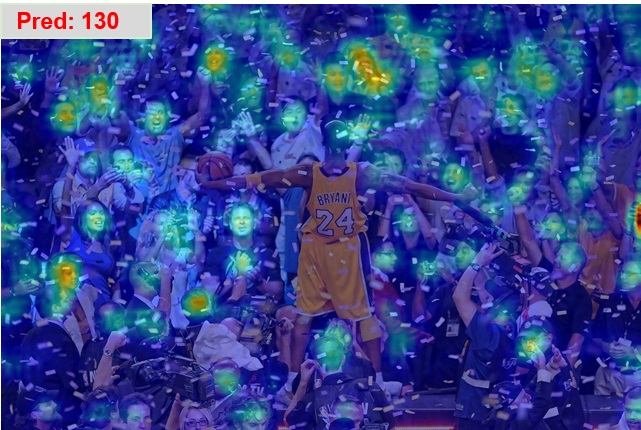} &
			\includegraphics[width=0.22\textwidth, height=31mm]{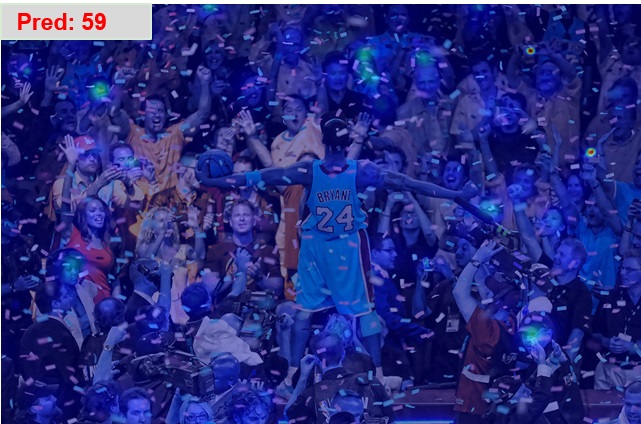} &
			\includegraphics[width=0.22\textwidth, height=31mm]{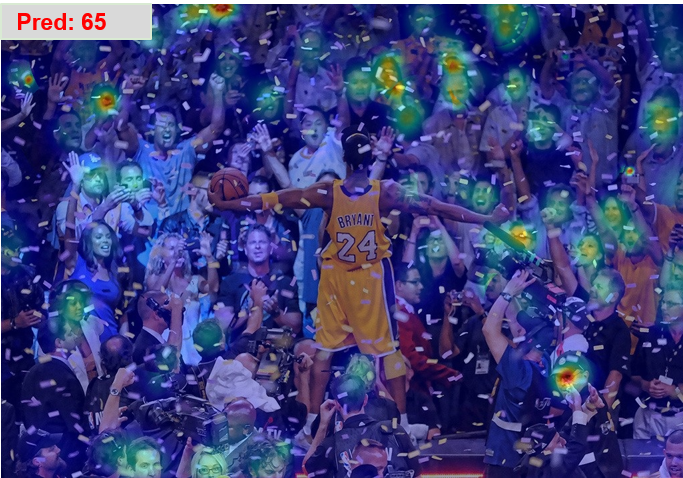} &
			\includegraphics[width=0.22\textwidth, height=31mm]{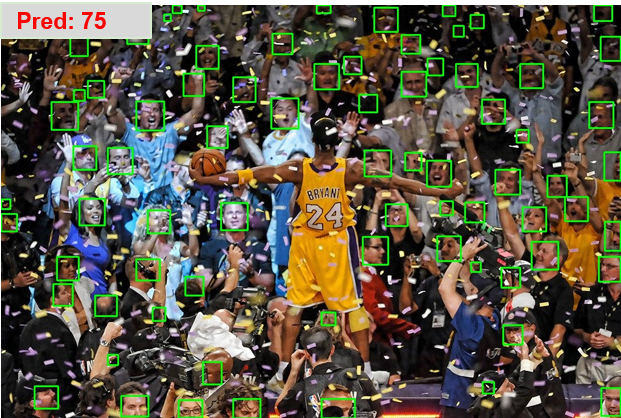} \\
			CSRNET~\cite{li2018csrnet} & DSSINET~\cite{liu2019crowd}  & Ours (density)  & Ours (localization) \\
	\end{tabular}
   \vspace{-3mm}
   \caption{Visualization results on some real scenes. We show the cross-scene results of CSRNET and DSSINET learnt on SHB as well as ours in the transfer setting of SHB $\rightarrow$ real scenes.  Our method produces better visual result and more accurate counting result over others.} 
\label{fig:real}
\end{figure*}

\subsection{Investigation on \textit{Reg-to-Det}}
The \textit{Reg-to-Det} transformer $\Phi$ is considered as a scene-agnostic operator in crowd counting. It is learnt in the source dataset and applied to the target dataset. This session evaluates the quality of $\Phi$ in the transfer setting A $\rightarrow$ Q and A $\rightarrow$ B by measuring the detection mAP of the transformed individual localization maps at iteration 0. 
Table~\ref{tab:transfer_knowledge} presents the results of the proposed encoder-decoder with different loss functions. It shows that the proposed focal MSE loss $L_{Focal-MSE}$ demonstrates a strong superiority over the conventional MSE loss $L_{MSE}$ (\eg 0.347 \vs 0.423 on A $\rightarrow$ Q). Meanwhile, adding the DMS-SSIM loss further improves the result ($L_{\Phi}$) to the best (\eg 0.448 on A $\rightarrow$ Q). This justifies the usage of focal loss and DMS-SSIM loss properly. We have also evaluated a focal cross entropy (CE) loss in Table~\ref{tab:transfer_knowledge} for another comparison: with $L_{Focal-CE}$, we achieve mAP 0.399 and 0.554 on A $\rightarrow$ B and A $\rightarrow$ Q, respectively; they are lower than using $L_{Focal-MSE}$.

Another interesting thing worth to mention is, it seems to be sufficient to learn the encoder-decoder ($\Phi$) with limited data in the source. For instance, with 30\% data, the mAP for A $\rightarrow$ Q and A $\rightarrow$ B has already reached 0.406 and 0.610, \vs 0.448 and 0.613 with 100\% data. We believe this is another evidence to prove the \textit{Reg-to-Det} transformer is a scene-agnostic operator: learning it from a small amount of data should be sufficient enough to achieve a reliable solution generalized over a large amount of data.

\tabcolsep=10pt
\begin{table}[t] 
\caption{Investigation on the \emph{Reg-to-Det} transformer in the transfer setting A $\rightarrow$ Q and A $\rightarrow$ B. AP is reported for individual localization performance.}
\vspace{-3mm}
\centering
\label{tab:transfer_knowledge}
\resizebox{1\linewidth}{!}{
\begin{tabular}{l|l|cc}
\toprule
  Solution  & Loss &  {\textbf{A$\rightarrow$Q}} &  {\textbf{A$\rightarrow$B}}\\
            \midrule
   \multirow{3}{*}{Encoder-decoder}   & $L_{MSE}$            &0.347 &0.482  \\
                & $L_{Focal-MSE}$         &0.423 &0.599  \\
                & $L_{Focal-CE}$         &0.399 &0.554  \\
                & $L_{\Phi}$              &0.448 &0.613   \\
\bottomrule
\end{tabular}}
\end{table}

\section{Conclusion}
In this paper, we propose an unsupervised crowd counting scheme via regression-detection {bi-knowledge transferring} from a labeled source dataset to an unlabeled target set. The mutual transformations between the output of regression and detection models are first investigated, which enables the mutual knowledge distillation between the two models in the target set. The regression and detection models are co-trained via a self-supervised way using their fused pseudo ground truth in the target set. Extensive experiments demonstrate the superiority of our method in the cross-domain setting for unsupervised crowd counting.

\begin{acks}
The research was partly supported by National Key Research and Development Program of China (2017YFB0802300), National Natural Science Foundation of China (No.61828602, No.61773270, \& No.61971005), JST CREST Grant (JPMJCR1686), and Grant-in-Aid for JSPS Fellows (18F18378).
\end{acks}

\bibliographystyle{ACM-Reference-Format}
\bibliography{egbib}

\end{document}